\documentclass[twoside]{article}

\usepackage{fullpage} 

\usepackage{palatino, epsfig, latexsym, natbib, algorithmic, epstopdf, amsmath, amssymb, color, amsthm, enumitem, graphicx, hyperref}

\newtheorem{definition}{Definition}

\newcommand{\ignore}[1]{}

\begin{document}

\title{\bf Evolutionary computation for multicomponent problems: opportunities and future directions}  

\author{
 Mohammad Reza Bonyadi \\
 mrbonyadi@cs.adelaide.edu.au, rezabny@gmail.com\\ 
 Department of Computer Science, The University of Adelaide, 
			Adelaide, 5005, Australia, \\ as well as Centre for Advanced Imaging, University of Queensland, QLD 4072, Australia 
		\and
 Zbigniew Michalewicz \\
zbyszek@cs.adelaide.edu.au\\
Department of Computer Science, The University of Adelaide, 
		Adelaide, 5005, Australia. \\ Also with Institute of Computer Science, Polish Academy of Sciences, Warsaw, Poland, \\ Polish-Japanese Institute of Information Technology, Warsaw, Poland, \\ and Complexica, Adelaide, Australia
	\and
 Frank Neumann \\
 frank.neumann@adelaide.edu.au\\ 
 Department of Computer Science, The University of Adelaide, 
		Adelaide, 5005, Australia
		\and
 Markus Wagner \\
 markus.wagner@adelaide.edu.au\\ 
 Department of Computer Science, The University of Adelaide, 
		Adelaide, 5005, Australia
}	
	
	\maketitle

\begin{abstract}
Over the past 30 years many researchers in the field of evolutionary computation have put a lot of effort to introduce various approaches for solving hard problems. Most of these problems have been inspired by major industries so that solving them, by providing either optimal or near optimal solution, was of major significance. Indeed, this was a very promising trajectory as advances in these problem-solving approaches could result in adding values to major industries. In this paper we revisit this trajectory to find out whether the attempts that started three decades ago are still aligned with the same goal, as complexities of real-world problems increased significantly. We present some examples of modern real-world problems, discuss why they might be difficult to solve, and whether there is any mismatch between these examples and the problems that are investigated in the evolutionary computation area.
\end{abstract}

\section{Motivation}
\label{sec:motivation}
The Evolutionary Computation (EC) community over the last 30 years has made a lot of effort designing optimization methods (specifically Evolutionary Algorithms, EAs) that are well-suited for hard problems --- problems where other methods usually fail~\citep{michalewicz2012quovadis}. As most real-world problems\footnote{See~\citep{michalewicz2012quovadis} for details on different interpretations of the term ``real-world problems''.} are quite complex, set in dynamic environments, with nonlinearities and discontinuities, with variety of constraints and business rules, with a few, possibly conflicting, objectives, with noise and uncertainty, it seems there is a great opportunity for EAs to be applied to such problems.

Some researchers investigated features of real-world problems that served as ``reasons'' for difficulties that EAs experience in solving them. For example, in~\citep{weise2009why} the authors discussed premature convergence, ruggedness, causality, deceptiveness, neutrality, epistasis, and robustness, that make optimization problems hard to solve. However, it seems that these reasons are either related to the landscape of the problem (such as ruggedness and deceptiveness) or the optimizer itself (like premature convergence and robustness) and they are not focusing on the nature of the problem. In~\citep{michalewicz2004howto}, a few different reasons behind the hardness of real-world problems were discussed; that included the size of the problem, presence of noise, multi-objectivity, and presence of constraints. Apart from these studies there have been EC conferences (e.g., GECCO, IEEE CEC, PPSN) that have had special sessions on ``real-world applications''. The aim of these sessions was to investigate potentials of EC methods in solving real-world optimization problems. 

Most of the features discussed in the previous paragraph have been captured in optimization benchmark problems (many of these benchmark problems can be found in OR-library\footnote{Available at: \url{http://people.brunel.ac.uk/~mastjjb/jeb/info.html}}). As an example, the size of benchmark problems has been increased during the last decades and new benchmarks with larger problems have appeared (e.g., knapsack problems, KP, with 2,500 items and traveling salesman problems, TSP, with more than 10,000 cities). Presence of constraints has also been captured in benchmark problems (e.g., constrained vehicle routing problem, CVRP) and studied by many researchers. Some researchers also studied the performance of evolutionary optimization algorithms in dynamic environments \citep{nguyen2012Continuous,jin2005uncertain}. Thus, the expectation is, after capturing all (or at least some) of these pitfalls and addressing them, EC optimization methods should be effective in solving real-world problems. However, after over 30 years of research and many articles written on Evolutionary Algorithms in dedicated conferences and journals with special sessions on applications of evolutionary methods on real-world applications, still it is not that easy to find EC-based applications in real-world. 

There are several reasons~\citep{michalewicz2012emperor} for such mismatch between the contributions made by researchers to the field of Evolutionary Computation over many years and the number of real-world applications which are based on concepts of Evolutionary Algorithms. These reasons include:

\begin{enumerate}[label=(\alph*)]
\item Experiments focus on single component (also known as single silo) benchmark problems
\item Dominance of Operation Research methods in industry
\item Experiments focus on global optima
\item Theory does not support practice
\item General dislike of business issues in research community
\item Limited number of EA-based companies
\end{enumerate}

It seems that the reasons (a) and (b) are primary while the reasons (c) to (f) are secondary. Let us explain.

There are thousands of research articles addressing traveling salesman problems, job shop and other scheduling problems, transportation problems, inventory problems, stock cutting problems, packing problems, various logistic problems, to name but a few. Although most of these problems are NP-hard and deserve research efforts, they are not exactly what the real-world industries need. Most companies run complex operations and they need solutions for problems of high complexity with several components (i.e., multicomponent problems\footnote{There are concepts similar to multicomponent problems in other disciplines, e.g., OR and management sciences, with different names such as integrated system, integrated supply chain, system planning, and hierarchical production planning.}). In fact, problems in real-world usually involve several smaller subproblems (several components) that interact with each other and companies are after a solution for the whole problem that takes all components into account rather than only focusing on one of the components. For example, the issue of scheduling production lines (e.g., maximizing the efficiency or minimizing the cost) has direct relationship with inventory costs, transportation costs, delivery-in-full-on-time to customers, etc., hence it should not be considered in isolation. Moreover, optimizing one component of the operation may have negative impact on other activities in other components. These days businesses usually need ``global solutions'' for their operations that includes all components together, not single-component solutions. This was recognized already over 30 years ago by Operations Research (OR) community; in~\citep{ackoff1979future} there is a clear statement: ``Problems require holistic treatment. They cannot be treated effectively by decomposing them analytically into separate problems to which optimal solutions are sought.'' However, there are very few research efforts which aim in that direction that is mainly due to the lack of appropriate benchmarks or test cases available. It is usually harder to work with a company on such global level because the delivery of a successful software solution involves many other (apart from optimization) skills such as understanding the company's internal processes and complex software engineering issues.

Further, there are many reasons why OR methods are widely used to deal with such problems. One reason is that the basic OR approaches (e.g., linear programming) are introduced to many students in different disciplines. This makes these approaches well-known by researchers and industries, and, consequently, widely used. Also, OR community has a few standard and powerful configurable products (e.g., CPLEX) that are used in many organizations especially for complex systems with many components~\citep{Kanban2006Wang} (see also \citep{michalewicz2012emperor} for further discussion).  

\begin{figure}
	\centering
	\centering
	\begin{tabular}{cc}
			\includegraphics[clip,height=40mm,width=40mm]{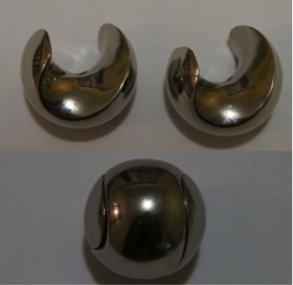} & \includegraphics[clip,height=40mm,width=40mm]{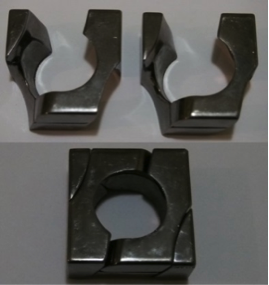}\\
			(a) &(b)\\
			\includegraphics[clip,height=40mm,width=40mm]{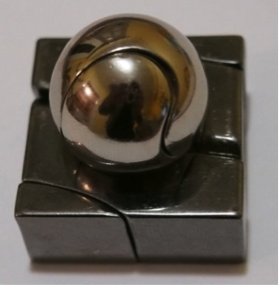}& \includegraphics[clip,height=40mm,width=40mm]{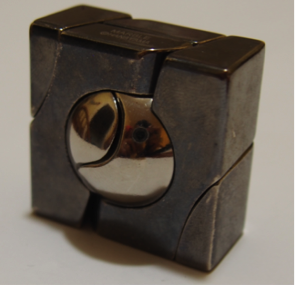}\\	
			(c) & (d)\\
	\end{tabular}
\caption{The Cast Marble puzzle: (a) two pieces of a ball and the ball that is generated by setting up the pieces, (b) two pieces of a cuboid and the cuboid that is generated by setting up the pieces, (c) the ball is not fit in the hole of the cuboid if the cuboid is set up first, and (d) the solution of the puzzle.}
	\label{fig:puzzle}
\end{figure}

Let us illustrate differences between single-component and multicomponent problems by presenting a puzzle (see Figure \ref{fig:puzzle})\footnote{The name of this puzzle is the ``Cast Marble'', created by Hanayama company.}. There is a ball that has been cut into two parts in a special way, and a cuboid with a hole inside, that has been also cut into two parts in a special way. The two parts of the ball can be easily set up together to make a complete ball (see Figure~\ref{fig:puzzle}a). Also, the two parts of the cuboid can be put easily together to shape the cuboid (see Figure~\ref{fig:puzzle}b). The size of the hole inside the cuboid is slightly larger than the size of the ball, so that, if the ball is inside the hole it can spin freely. However, it is not possible to set up the cuboid and then put the ball inside the cuboid as the entry of the hole of the cuboid is smaller than the size of the ball (see Figure~\ref{fig:puzzle}c). Now, the puzzle is stated as follows: set up the cuboid with the ball inside (see figure Figure~\ref{fig:puzzle}d). Setting up the ball separately and the cuboid separately is easy. However, setting up the cuboid while the ball is set up inside the cuboid is extremely hard.\footnote{The difficulty level of this puzzle was reported as 4 out of 6 by the Hanayama website, that is equal to the difficulty of Rubik’s cube.} 

This puzzle nicely represents the difference between single-component and multicomponent problems. In fact, solving a single component problem (setting up the ball or the cuboid separately) might be easy; however, solving the combination of two simple component problems (setting up the ball while it is inside the cuboid) is potentially extremely harder. 

The purpose of this letter is to encourage the EC community to put more effort in researching multicomponent problems. First, OR community is already doing this. Second, such research seems necessary if we would like to see the emergence of powerful EC-based applications in the real-world. Third, because of the flexibility of EC techniques, we believe that they are more than suitable for delivering quality solutions to multicomponent real-world problems.

So, in this letter we explore this issue further and we organize the letter as follows. In section \ref{sec:twoexamples} two real-world examples are explained, in section \ref{sec:lessonslearned} some important observations about real-world problems are discussed, in section \ref{sec:TTP} a recently presented benchmark multicomponent problem is introduced and discussed, and in section \ref{sec:discussion} some discussions and directions for future research are provided.

\section{Two Examples}
\label{sec:twoexamples}
The first example relates to optimization of the transportation of water tanks~\citep{stolk2013watertanks}. An Australian company produces water tanks with different sizes based on some orders coming from its customers. The number of customers per month is approximately 10,000; these customers are in different locations, called stations. Each customer orders a water tank with specific characteristics (including size) and expects to receive it within a period of time (usually within one month). These water tanks are carried to the stations for delivery by a fleet of trucks that is operated by the water tank company. These trucks have different characteristics and some of them are equipped with trailers. The company proceeds in the following way. A subset of orders is selected and assigned to a truck and the delivery is scheduled in a limited period of time (it is called subset selection procedure). Because the tanks are empty and of different sizes they might be packed inside each other (it is called bundling procedure) to maximize truck’s load in a trip. A bundled tank must be unbundled at special sites, called bases, before the tank delivery to stations. Note that there might exist several bases close to the stations where the tanks are going to be delivered and selecting different bases (it is called base selection procedure) affects the best overall achievable solution. When the tanks are unbundled at a base, only some of them fit in the truck as they require more space. The truck is loaded with a subset of these tanks and carries them to their corresponding stations for delivery. The remaining tanks are kept in the base until the truck gets back and loads them again to continue the delivery process (it is called delivery routing procedure). 

The aim of the optimizer is to divide all tanks ordered by customers into subsets that are bundled and loaded in trucks (possibly with trailers) for delivery and to determine an exact routing for bases and stations for unbundling and delivery activities – to maximize the total ``value'' of the delivery at the end of the time period. This total value is proportional to the ratio between the total prices of delivered tanks to the total distance that the truck travels. 

Each of the mentioned procedures in the tank delivery problem (subset selection, base selection, and delivery routing, and bundling) is just one component of the problem and finding a solution for each component in isolation does not lead to the optimal solution of the whole problem. As an example, if the subset selection of the orders is solved to optimality (the best subset of tanks is selected in a way that the price of the tanks for delivery is maximized), there is no guarantee that there exists a feasible bundling such that this subset fits in a truck. Also, by selecting tanks without considering the location of stations and bases, the best achievable solutions might not be very high quality, e.g., there might be a station that needs a very expensive tank but it is very far from the base, which actually makes delivery very costly. On the other hand, it is impossible to select the best routing for stations before selecting tanks---without selection of tanks, the best solution (lowest possible tour distance) is to deliver nothing. Thus, solving each subproblem in isolation does not necessarily lead to the overall optimal solution.

Note also that in this particular case there are many additional considerations that must be taken into account for any successful application. These include scheduling of drivers (who often have different qualifications), fatigue factors and labor laws, traffic patterns on the roads, feasibility of trucks for particular segments of roads, maintenance schedule of the trucks. 

The second example relates to the optimization of a wine supply chain \citep{michalewicz2010optimising}, from grape to bottle. The overall aim of the wine producer is to deliver the orders in time while minimize the useless storage. The wine producer needs to decide if the grape is at its peak of ripeness, i.e., optimal maturity, to be collected and used for wine production. This is done through a predictive model that assesses different characteristics of the grape (e.g., sugar, acidity, berry metabolites, berry proteins, taste) to determine when is the grape in its ''optimal maturity'' for harvest. The definition of optimal maturity may vary depending upon the style of wine being made; the working definition of quality; varietal; rootstock; site; interaction of varietal, rootstock and site; seasonal specific factors; viticultural practices; and downstream processing events and goals. Once the ripeness was verified, the grapes are removed and sent to the weighbridge (this stage is called intake planning). After weighting the grapes, they are crushed using specific crushers to provide grape juice (this is called crusher scheduling). The grape juice is then stored in some tanks where they are fermented to provide wine, each tank may have different capacities and capabilities (this is called storage scheduling). Different types of wine may require specific tank attributes for processing, such as refrigeration jackets or agitators. As some special types of wine may need a blend of different grapes juice, it is better to store the juice in adjacent tanks to facilitate such blending if necessary. Also, it is better to fill/use the tanks to their full capacity as a half-empty tank affects the quality of the wine inside to a degree that the wine might become useless (because of its quality) after a while. To prepare the final product, the wines in the tanks should be bottled that requires scheduling the bottling lines (this is called the bottling line scheduling). This is done through bottling lines where appropriate tank of wine (according to the placed orders) are connected to the line and the wine is bottled and sent for either storage or to direct delivery. 

The aim of the optimizer is to find a feasible schedule for the intake plan to remove grapes in their pick of ripeness, schedule crushers to press these grapes, assign the crushed grapes to proper tank farms with appropriate facilities depending on the desired wine, and schedule bottling lines to the tank farms to perform bottling and deliver placed orders as soon as possible. 

Each of the mentioned models in the optimization of the wine supply chain (intake planning, crushing stage, storage scheduling, and bottling line scheduling) is just one component of the overall problem and finding a solution for each component in isolation does not lead to the optimal solution of the whole problem. For example, daily decisions on the crushing should not be done in isolation from storage of wines and juices in the tank farm, as even if the crushing can be done in a very efficient way, the capacity of tanks might constraint the process. Also, the storage scheduling should consider the crushing too as the optimal choice for the storage depends on the amount and type of the processed material in the crushing stage. 

In the real-world case, there are some other considerations in the problem such as scheduling of workers for removing grapes and required transports, maintenance schedule for tools and machines, deal with sudden changes and uncertainty (e.g., weather forecast) and take into account risk factors.

\section{Lessons Learned - Dependencies}
\label{sec:lessonslearned}

Let us take a closer look at the examples presented in section \ref{sec:twoexamples}. Obviously, both optimization problems contain constraints and noise; both of them might be large in terms of the number of decision variables. However, there is another characteristic present in both problems: each problem is a combination of several subproblems (components/silos). The tank delivery problem is a combination of tank selection, delivery routing, base selection, and bundling. Also, the wine industry problem is a combination of intake planing, crushers scheduling, storage scheduling, and bottling line scheduling. Because of this characteristic we call them multicomponent problems. Each component might be hard to solve in isolation and solving each component to optimality does not necessarily direct the search towards good overall solutions if other components are not considered. In fact, solutions for each subproblem potentially affect the variables of some other subproblems because of the dependencies among components. As an example, in the tank delivery problem, delivery routing (best route to deliver tanks) is affected by the base selection (best choice for the base to unbundle the tanks) as choosing different bases imposes different lengths of travels between the base and stations. Delivery routing is also affected by the tank selection (selecting the tanks for delivery) as the selected subset of tanks determines the stations to visit. In short, a solution for each component affects the feasibility or the best achievable solution in other components. In spite of the importance of this topic in real-world problems, the progress to address such problems with dependencies has been very limited in the EC community so far.

Dependency causes appearance of some other features in real-world problems. As an example, the dependency among components in a problem sometimes causes a special flow of data among components of that problem. For instance, generating a solution for the delivery routing in the tank delivery problem is impossible without being aware of the selected subset of tanks. 

\begin{definition}\label{def:datafollower}
If generating a solution (independent from the quality) for a component $A$ is impossible because of some data that needs to be provided by the component $B$, we say that $A$ is the data follower of $B$. 
\end{definition}

This indeed imposes a special sequence of solution procedure for some components that need to be taken into account by the solver.

Also, dependencies among components make (mathematical) modeling of problems more complex. In fact, modeling of existing benchmark problems (such as TSP, multidimensional KP, job shop scheduling, etc) is relatively easy and many different models for them already exist. However, a multicomponent problem involves a more complex model, even if it has been composed of components that have been modeled in previous studies. The main reason is that in a multicomponent problem a constraint that is in one of the components may influence feasible set of solutions of other components because of dependency. Thus, modeling each component in isolation and putting these models together does not express the model for the whole problem. 

Another effect that is caused by dependency in multicomponent problems is the propagation of noise. Noise in benchmark problems is usually defined by a stochastic function like a normal or Poisson distribution that is simply added to the objective function or constraints. However, with the presence of dependency noise is propagated from each component to the others. Because dependency between components might follow a complex function, the propagated noise from one component to the others might also become complex (even if the original noise was based on a simple distribution) which causes difficulties in solving the problem. As an example, if the break-down distribution of the crushers in the wine supply chain problem is Poisson with some parameters and this break-down has some other distribution for the tanks maintenances, the effect of these noises on the objective function cannot be treated as a simple Poisson and it is actually hard to even estimate. Note also that investigation of noise over the whole system results in a better risk analysis in the whole operation and a better estimation tolerance in the final benefit.

An additional interesting feature that comes into play because of dependency is the concept of bottleneck component. A bottleneck component is a component in the whole system that constrains the best overall achievable solution. If there is a bottleneck component in the system, adding more resources to other components does not cause improvement of the best achievable objective value (see also~\citep{bonyadi2014bottlenecks} for details). As an example, in the wine supply chain problem if the bottleneck component is the number of crushers then investment on any other parts of the system (expanding the number of tanks, hiring more workers to remove grapes, or establishing new bottling line) has minimal affect (or even no effect) on the best overall achievable solution in the system.

Dependency is the source of some complexities in multicomponent problems. Thus, a proper definition for dependency among components is of importance. The concept of dependency among components and effects of the components on each other is similar to the concept of separability of variables in continuous space optimization problems~\citep{whitley1997island}. As it was mentioned earlier, dependency stems from the effects of components on each other, i.e., changing the solution of a component affects feasibility or quality of solutions of other components. Accordingly, we suggest the following definition for dependency among components:

\begin{definition}\label{def:dependency}
We say component $B$ is dependent on component $A$ (notation: $A \rightarrow B$) if \\
(1) $A$ is not data follower of $B$ (see Definition~\ref{def:datafollower}),
and \\
(2) changing the solutions of component $A$ can change the best achievable solution for the component $B$ in terms of the overall objective value. 
\end{definition}

\begin{figure}
\centering
\centering
\includegraphics[clip,width=60mm]{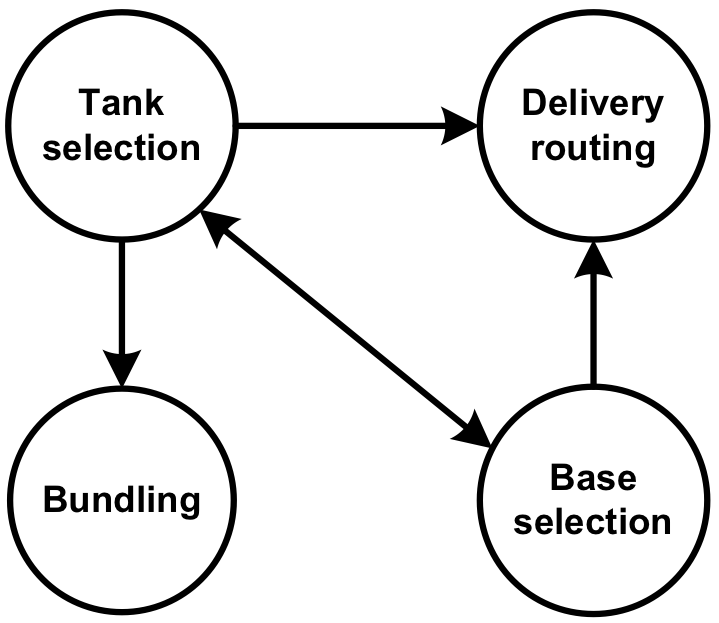}
\caption{Diagram of dependency among components of tank delivery problem.}
\label{fig:dependencies}
\end{figure}

The part (1) of the definition prevents introduction of dependency between two components, $A$ and $B$, where $A$ needs a flow of data from $B$. If $A$ is a data follower of $B$ then $B$ cannot be dependent to $A$. Assessing the first part of the definition is not a hard task. \ignore{As an example, generating a solution for the train scheduling in the mine to port problem is possible without being aware of unload schedule. However, it is impossible to generate a solution for the unload schedule without knowing the solution for the train schedule.} The part (2) of the definition ensures that the components are not separable. To assess if $B$ is dependent to $A$, assume that there exist a solution $ a $ for the component $A$ and, given $A$ is fixed to $ a $ (showing by $A=a$), setting $B=b$ results in the best possible overall objective value. Now, if there exists an $a’$ that for $A=a’$, $B=b'\neq b$ results in the best overall objective value, then $A\rightarrow B$ ($B$ is dependent to $A$). This means that, changing the solution for $A$ actually might change the best solution for $B$. Dependency is shown in a diagram for the example problems (see Figure~\ref{fig:dependencies}).

Figure~\ref{fig:dependencies} shows the diagram of dependency among components of the tank delivery problem. The links in the figure refer to a dependency among different components. As an example, one can fix the solution for the base selection (of course with being aware of the solution for the subset selection) without being aware of the delivery routing solution. Also, by changing the base, the best achievable solution (shortest tours) for delivery routing is changed. Thus, the base selection is linked to delivery routing. Note that generating a solution for delivery routing is impossible without being aware of the location of the base, hence, there is no link from delivery routing to the base selection. 

Hypothetically, any dependency can exist between a set of problems. These dependencies can be represented by a digraph, which can potentially form a complex network. In the simplest case, there are some problems with no dependencies. In this case, one can solve each problem separately to optimality and combine the solutions to get the global optimal solution. 

\section{Traveling Thief Problem (TTP)}
\label{sec:TTP}
A recent attempt ~\citep{bonyadi2013ttp} to provide an abstraction of multicomponent problems with dependency among components was introduced recently; it was called the traveling thief problem (TTP). This abstraction combined two problems and generated a new problem, which contains two components. The TSP and KP were selected and combined, as both problems were well known and researched for many years in the field of optimization. TTP was defined as follows. A thief is supposed to steal $ m $ items from $ n $ cities, where the distances $d_{j,k}$ between cities $j$ and $k$, profits of each item ($p_i$), and weights of the items ($w_i$) are given. The thief carries a limited-capacity $W$ knapsack to store the collected items. The problem is to find the best plan for the thief (in terms of maximizing its total benefit) to visit all cities exactly once (a TSP component) and select the items from these cities (a KP component) so its total benefit is maximized. 

To make these two components dependent, it was assumed that the current speed of the thief is affected by the current weight of the knapsack ($W_c$). In other words, the more items the thief selects, the slower he can move. A function $v:R \rightarrow R$ was defined that maps the current weight of the knapsack to the current speed of the thief: $v(W)$ is the minimum speed of the thief (full knapsack) and $v(0)$ is the maximum speed of the thief (empty knapsack). Further, the thief’s profit is reduced (e.g., rent of the knapsack, $r$) by the total time he needs to complete the tour. So the total profit $B$ of the thief is then calculated by

$$B=P-r \times T$$

\noindent where $P$ is the aggregation of the profits of the selected items, $r$ is the rent of the knapsack, and $T$ is the total tour time.

\begin{figure}
	\centering
	\centering
	\includegraphics[clip,width=60mm]{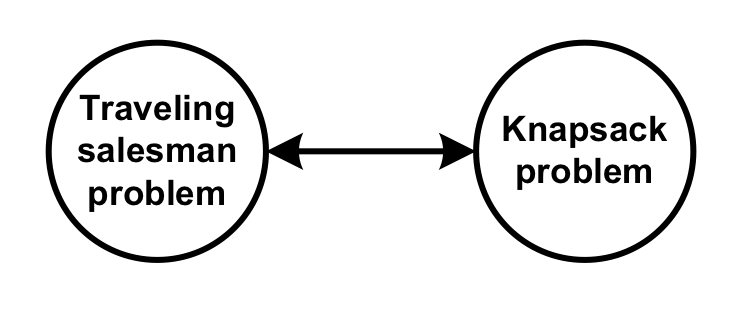}
	\caption{TTP dependency graph.}
	\label{fig:ttpDependencyGraph}
\end{figure}

It is possible to generate solutions for KP or TSP in TTP sequentially. Note, however, that each solution for the TSP component impacts the best possible solution for the KP because of the total profit is a function of travel time. Further, each solution for the KP component impacts the tour time for TSP as the weight of the knapsack impacts the speed of travel due to the variability of weights of items. 

Note that different values of the rent $r$ and different functions $v$ result in different instances of TTPs that might be ``harder'' or ``easier'' to solve. For example, for small (relative to $P$) values of $r$, $r \times T$ contributes a little to the value of $B$. In the extreme case ($r=0$), the contribution of $r \times T$ is zero, so the best solution for a given TTP is equivalent to the best solution of the KP component. In other words, in such a case there is no need to solve the TSP component at all. By increasing the value of $r$, the contribution of $r \times T$ becomes larger – and if the value of $ r $ is very large (relative to $P$) then the impact of $P$ on $B$ becomes negligible. In such a case the optimum solution of the TTP would be very close to the optimum solution of the given TSP. 

\newcommand{\specialExpression}{$ \left| (v(W)-v(0))/W \right| $} 
\renewcommand{\specialExpression}{$ \left| \frac{v(W)-v(0)}{W} \right| $} 

\begin{figure}
	\centering
	\centering
	
	\begin{tabular}{cc}
		\includegraphics[clip,width=0.4\textwidth]{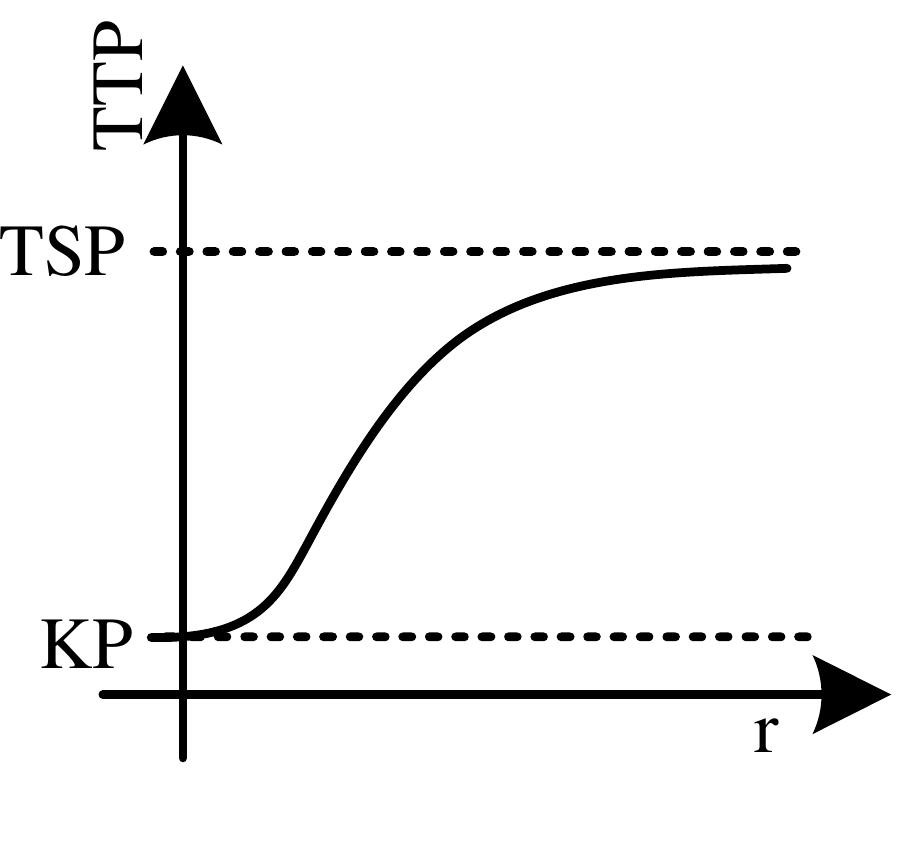} & \includegraphics[clip,width=0.4\textwidth]{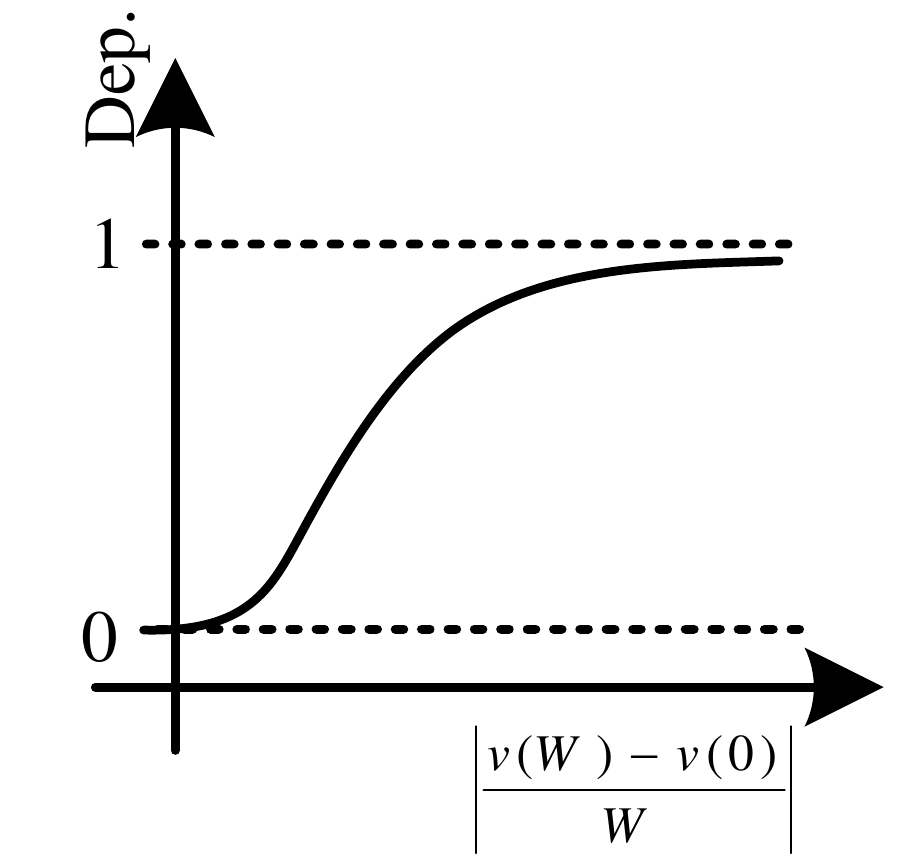}\\
		(a) & (b)\\
	\end{tabular}
	
	\caption{(a) The impact of the rent rate $r$ on dependency in TTP. For $r=0$, the TTP solution is equivalent to the solution of KP, while for larger $r$ the TTP solutions become more closer to the solutions of TSP. (b) The impact of the speed $ v $ on dependency in TTP. When $ v $ does not drop significantly for different weights of picked items (\specialExpression is small), the two problems can be decomposed and solved separately. The value Dep.=1 (Dep. is the short for Dependency) shows that the two components are dependent while Dep.=0 shows that two components are not dependent. }
	\label{fig:ttpRentRateImpact}
\end{figure}

Similar analysis can be done for the function $v$. For a given TSP and KP components, different functions $v$ would result in different instances of TTPs. These instances, as before, might be ``harder'' or ``easier'' to solve. For example, let’s assume that $v$ is a decreasing function of weight of the knapsack, i.e., selecting items with positive weight would not increase the value of $v$. In such a case, if selected items do not affect the speed of the travel (i.e., \specialExpression is zero) then the optimal solution of the TTP is the composition of the optimal solutions of KP and TSP that are solved separately, as selection of items does not change the time of the travel. As the value of \specialExpression grows, the TSP and KP become more dependent on each other (i.e., selection of items have more significant impact on the travel time), so selecting more items reduces the value of $B$ significantly. In the extreme case (\specialExpression is infinitely large) it would be better not to pick any item (i.e., the solution for the KP would be to pick no items at all) and only solve the TSP component as efficiently as possible. This has been also discussed in~\citep{bonyadi2014social}.

Recently other researchers reported some experimental results and analytical investigations of TTP. Opportunities to decompose instances of TTP to KP and TSP were studied in~\citep{mei2014investigation}. Reported experimental results indicated that methods which consider both components simultaneously consistently achieve better results than methods which decompose the problem and solve them separately. Further, some researchers \citep{polyakovskiy2014ttpbenchmark} proposed a comprehensive benchmark set for TTP and the experimental results on performances of three methods (one simple heuristic and two evolutionary based methods) were reported.\footnote{Code, results, and further articles are available on our project page: \url{cs.adelaide.edu.au/~optlog/research/ttp.php}} 
The large size TTP instances proposed in \citep{polyakovskiy2014ttpbenchmark} were further studied in ~\citep{mei2014improving} where the authors investigated an improved evolutionary algorithm to deal with large scale TTPs.

\section{Discussion and Future Directions}
\label{sec:discussion}

There are a few challenges in dealing with multicomponent problems. In this section we discuss some of these challenges and present some potential opportunities for EC-based methods to address them. 

As it was mentioned earlier, a collection of optimal solutions that correspond to components of a multicomponent problem does not guarantee global optimality. This is because of presence of dependencies among components. To solve a multicomponent problem, however, it is often necessary to decompose the problem taking into account the dependencies among components. Complexity of such decomposition is usually related to the dependencies among the components. For example, if in the cast marble puzzle the final objective was to set up the ball on top of the cuboid (a simple dependency between components) then the puzzle would be extremely easy (one could set up the ball and the cuboid separately and put the ball on the cuboid). In contrast, setting up the ball inside the cuboid (a complex dependency between components) is extremely hard as the ball and the cuboid need to be set up simultaneously one inside another. Likewise, one can define a simple dependency between KP and TSP in the TTP problem that makes the problems decomposable or make them tighter together so that they are not easily decomposable.

The lack of abstract problems that reflect the presence of dependencies among components is apparent in the current benchmarks\footnote{There are some problems such as multiprocessor task scheduling problem (MTSP) \citep{bonyadi2009bipartite} or resource investment scheduling problem (RISP)\citep{Abbas2014RIP} that can be also considered as two-component problems with a simple dependency between the components.}. In fact, real-world supply chain optimization problems are a combination of many smaller subproblems dependent on each other in a network while benchmark problems are singular. Because global optimality (in the sense of the overall objective) is of interest for multicomponent problems, singular benchmark problems cannot assess quality of methods which are going to be used for multicomponent real-world problems with the presence of dependencies. 

One of the challenges in solving multicomponent problems relates to the way the problem is modeled. There are two general ways to model such problems. One way is to design a monolithic model to represent all variables, constraints, and objectives of all components of the problem and then design a method to apply to such a model (see \citep{Kanban2006Wang} for example for a complex large linear model for a supply chain problem). The other way is to model the components separately and consider the dependencies to integrate the components. A monolithic model for a multicomponent problem carries some disadvantages. As an example, such a model is usually large and hard to define because of potential complications raising form different variables in different domains, different constraints in different components and their potential relations, etc. Also, it is hard to maintain such a model because all components have been fed into a large system (when for example a new component is added or a configuration is changed then the whole model should be redesigned). Finally, such a model usually disregards potential knowledge about the individual components. Note also that the model needs to represent the details of the system such as potential noises and dynamicity on different components that might be of different natures. These issues are, however, easier to address if the model is composed of smaller models, each model represent a component, and their integration represents a modeling for the problem, i.e., a multicomponent model. Such a model is easier to define as it follows the natural representation of the problem, is easier to maintain because of its modularity (adding, removing, or modifying a component only affects a part of the whole model), and enables the possible usage of existing knowledge, algorithms, and modeling ideas (with modifications) to deal with the problem as some components might be already well-studied\footnote{An example of different modeling can be seen for MTSP: one can design a large model to solve MTSP \citep{moghaddam2012immune} or, alternatively, the components can be modeled separately and different methods are applied to the components and the results are integrated to solve the problem \citep{bonyadi2009bipartite}.}. 
	
One can also consider modeling a multicomponent problem in several levels~\citep{DebBilelvelBase} and then apply the bi-level optimization (and in the more general case, the multi-level optimization) approaches to that model of the problem. This approach for modeling is in fact a special case of multicomponent modeling. There have been some advances related to single-objective~\citep{LegillonLT12} and multi-objective~\citep{DebBilevel} bi-level optimization in the evolutionary computation community. Bi-level models of a problem can be split into an upper and a lower level that depend on each other. Usually, there is a leader-follower relation between the upper and lower level. For the multi-component problems, however, such relation might not exist. Instead, we can have multiple problems that depend on each other in an arbitrary way. As a bi-level problem can be seen as a special case of a multi-component problem, we assume that techniques developed in the area of evolutionary bi-level optimization can be very useful for designing evolutionary algorithms for multi-component problems.

Clearly, choosing among different modeling approaches for multicomponent problems depends on the problem at hand. However, there might be situations that existing knowledge about the problem and its components can assist the designer to model the problem. As an example, if there is existing knowledge about the components of a multicomponent problem (as some components are well-studied problems), then it might be better to use the multicomponent modeling (rather than the monolithic modeling) approach – as the existing knowledge may assist in solving the problem effectively. Also, if there is a well-studied bi-level model that can formulate the multicomponent problem at hand, then it might be better to use that model to make use of existing knowledge about the model. Nevertheless, there might be a need to tailor the existing models to fit the components of the problem at hand. Such tailoring should be conducted with consideration of the dependencies as, otherwise, the solution of the model might deviate (potentially significantly) from the solution of the original problem. In addition, it might be possible to model the problem at hand as a single objective or multiple objectives. This again depends on the problem and potential existing knowledge about the objectives as well as the nature of the objectives (whether they can be combined into one objective or, in the case of single objective, if that objective can be decomposed into several objectives). One should note that combining objectives might result in irregularities in the landscape of the problem and make the problem harder or easier to solve for different optimization algorithms.

Typically EC methods offer a great flexibility in terms of incorporating several factors (constraints, multiple objectives, noise, etc.) that allows a designer to retain intricacies of the problem in the model. One possible EC-based approach to deal with a multicomponent model involves cooperative coevolution: a type of multi-population Evolutionary Algorithm~\citep{potter1994cooperativecoevolution}. Coevolution is a simultaneous evolution of several genetically isolated subpopulations of individuals that exist in a common ecosystem. Cooperative coevolution uses divide and conquer strategy: all parts of the problem evolve separately; fitness of each individual of particular species is assigned based on the degree of collaboration with individuals of other species. It seems that cooperative coevolution is a natural fit for multicomponent problems with presence of dependencies. Individuals in each subpopulation may correspond to potential solutions for particular components, with their own evaluation functions, whereas the global evaluation function would also include dependencies between components. In a more generic framework, one can consider a network of solvers (including heuristics, meta heuristics, and exact methods) to deal with each component where each solver decides individually while communicates with others to meet a global goal. 

Cooperative coevolution has already been used to address a few real-world multicomponent problems. For example, in \citep{ibrahimov2012part1,ibrahimov2012part2} a problem similar to the tank delivery example discussed in section \ref{sec:twoexamples} was formulated and a method based on cooperative coevolution was experimented with. Results of experiments showed that the cooperative coevolution method can provide high quality solutions. These applications were, however, developed ``ad hoc'' with the lack of any guidance from theoretical investigations. One can consider a systematic way to investigate the dependencies among components and their potential impacts on the performance of the algorithm and its parameters. Also, investigation of different approaches to deal with different components and integration of the solutions can be other research topics in this regard.

The method described in \citep{ibrahimov2012part1,ibrahimov2012part2} could also provide the solutions within a 10 minutes time frame, that is another important aspect that need to be satisfied in solving real-world problems. The reason is that decisions in real-world problems need to be taken in a small amount of time (5-10 minutes), otherwise, those decisions are not of much use. A team of OR experts designed a method to solve a multicomponent problem with almost the same complexity of the one considered in \citep{ibrahimov2012part1,ibrahimov2012part2}. After application of the algorithm to the real-case it was found that the algorithm needs 18 hours per objective to deal with the problem\footnote{Private correspondence.}. Such delay for providing a solutions is, however, not acceptable. 

One should note that the concept of global optimality is usually not of great importance when dealing with real-world problems. One reason is that the model that is designed for the problem (whether it is a monolithic model or several small models that need to be integrated to represent the problem) usually includes some simplifications. These simplifications can range from relaxations of constraints to accuracy adjustments of time-consuming exact simulators and to the creation of approximative surrogate models. Hence, as the solver is applied to this model, even if it solves the model to optimality, it still has some deviations from real-world (expected) solution because of the simplification. Also, because of the dynamic nature of the problems in real-world, even if the solver finds the optimum solution, that solution might not be valid anymore after a (possibly short) period of time due to unexpected events, e.g., delay of products, failure of trucks, extreme changes of whether (see also \citep{michalewicz2012quovadis, michalewicz2012emperor} for further discussion).
	
Multicomponent problems pose new challenges for the theoretical investigations of evolutionary computation methods. The computational complexity analysis of evolutionary computation is playing a major role in this field~\citep{auger2011theorybook,neumann2010theorybook}. Results have been obtained for many NP-hard combinatorial optimization problems from the areas of covering, cutting, scheduling, and packing. We expect that the computational complexity analysis can provide new rigorous insights into the interactions between different components of multicomponent problems. As an example, we consider again the TTP problem. Computational complexity results for the two underlying problems (KP and TSP) have been obtained in recent years. Building on these results, the computational complexity analysis can help to understand when the interactions between KP and TSP make the optimization process harder.

In a similar way, feature-based analysis might be helpful to provide new insights and help in the design of better algorithms for multicomponent problems. Analyzing statistical feature of classical combinatorial optimization problems and their relation to problem difficulty has gained an increasing attention in recent years~\citep{smithmiles2014towardsobjectivemeasures}. Classical algorithms for the TSP and their success depending on features of the given input have been studied in~\citep{smithmiles2010understandingtsp,mersmann2013novelapproach} and similar analysis can be carried out for the knapsack problem. Furthermore, there are different problem classes of the knapsack problem which differ in their hardness for popular algorithms~\citep{martello1990knapsack}. Understanding the features of the underlying subproblems and how the features of interactions in a multicomponent problem determines the success of different algorithms is an interesting topic for future research which would guide the development and selection of good algorithms for multicomponent problems.

It seems multicomponent problems provide great opportunity for further research in EC community. Thus, we believe that future research in this direction can potentially close the gap between academic research in EC community and needs for optimization methodologies in industries. 


\bibliographystyle{apalike}

\bibliography{references}
%

\end{document}